\PassOptionsToPackage{unicode}{hyperref}
\PassOptionsToPackage{hyphens}{url}
\documentclass{article}

\usepackage{amsmath,amssymb}
\usepackage{xcolor}

\usepackage{iftex}
\ifPDFTeX
  \usepackage[T1]{fontenc}
  \usepackage[utf8]{inputenc}
  \usepackage{textcomp}
  \usepackage{lmodern} 
\else
  \usepackage{unicode-math} 
  \defaultfontfeatures{Scale=MatchLowercase}
  \defaultfontfeatures[\rmfamily]{Ligatures=TeX,Scale=1}
\fi

\IfFileExists{microtype.sty}{\usepackage[]{microtype}
  \UseMicrotypeSet[protrusion]{basicmath}}{}

\makeatletter
\@ifundefined{KOMAClassName}{
  \IfFileExists{parskip.sty}{\usepackage{parskip}}{
    \setlength{\parindent}{0pt}
    \setlength{\parskip}{6pt plus 2pt minus 1pt}}
}{\KOMAoptions{parskip=half}}
\makeatother

\usepackage{longtable,booktabs,array,calc}
\usepackage{etoolbox}
\makeatletter
\patchcmd\longtable{\par}{\if@noskipsec\mbox{}\fi\par}{}{}
\makeatother
\IfFileExists{footnotehyper.sty}{\usepackage{footnotehyper}}{\usepackage{footnote}}
\makesavenoteenv{longtable}
\usepackage{graphicx}
\makeatletter
\newsavebox\pandoc@box
\newcommand*\pandocbounded[1]{%
  \sbox\pandoc@box{#1}%
  \Gscale@div\@tempa{\textheight}{\dimexpr\ht\pandoc@box+\dp\pandoc@box\relax}%
  \Gscale@div\@tempb{\linewidth}{\wd\pandoc@box}%
  \ifdim\@tempb\p@<\@tempa\p@\let\@tempa\@tempb\fi
  \ifdim\@tempa\p@<\p@\scalebox{\@tempa}{\usebox\pandoc@box}%
  \else\usebox{\pandoc@box}\fi}
\def\fps@figure{htbp}
\makeatother

\ifPDFTeX
  \newcommand{\RL}[1]{\beginR #1\endR}
  \newcommand{\LR}[1]{\beginL #1\endL}
  \newenvironment{RTL}{\beginR}{\endR}
  \newenvironment{LTR}{\beginL}{\endL}
\fi
\ifluatex
  \newcommand{\RL}[1]{\bgroup\textdir TRT#1\egroup}
  \newcommand{\LR}[1]{\bgroup\textdir TLT#1\egroup}

\fi

\IfFileExists{xurl.sty}{\usepackage{xurl}}{}
\usepackage{hyperref}
\hypersetup{hidelinks,pdfcreator={LaTeX via pandoc}}
\usepackage{bookmark}

\usepackage{authblk}
\makeatletter
\renewcommand\AB@affilsepx{ \par }  
\makeatother

\title{\textbf{Simulating a Bias Mitigation Scenario in Large Language Models}}

\title{\textbf{Simulating a Bias Mitigation Scenario in Large Language Models}}

\author[a]{Kiana Kiashemshaki}
\author[b]{Mohammad Jalili Torkamani}
\author[c]{Negin Mahmoudi}
\author[d,e]{Meysam Shirdel Bilehsavar}

\affil[a]{Department of Computer Science, Bowling Green State University, Bowling Green, Ohio, USA}
\affil[b]{School of Computing, University of Nebraska--Lincoln, Lincoln, Nebraska, USA}
\affil[c]{Department of Civil, Environmental, and Ocean Engineering, Stevens Institute of Technology, New Jersey, USA}
\affil[d]{Department of Computer Science, University of South Carolina, USA}
\affil[e]{Artificial Intelligence Institute, University of South Carolina, USA}

\affil[]{\textit{Corresponding author:} \href{mailto:kkiana@bgsu.edu}{kkiana@bgsu.edu}}

\affil[]{\textit{Contributing authors’ emails: }
\href{mailto:mJaliliTorkamani2@huskers.unl.edu}{mJaliliTorkamani2@huskers.unl.edu},
\href{mailto:nmahmoud1@stevens.edu}{nmahmoud1@stevens.edu},
\href{mailto:Meysam@email.sc.edu}{Meysam@email.sc.edu}}

\date{} 

\begin{document}
\maketitle

\section{Abstract:}\label{abstract}

Large Language Models (LLMs) have fundamentally transformed the field of
natural language processing; however, their vulnerability to biases
presents a notable obstacle that threatens both fairness and trust. This
review offers an extensive analysis of the bias landscape in LLMs,
tracing its roots and expressions across various NLP tasks. Biases are
classified into implicit and explicit types, with particular attention
given to their emergence from data sources, architectural designs, and
contextual deployments. This study advances beyond theoretical analysis
by implementing a simulation framework designed to evaluate bias
mitigation strategies in practice. The framework integrates multiple
approaches including data curation, debiasing during model training, and
post-hoc output calibration and assesses their impact in controlled
experimental settings. In summary, this work not only synthesizes
existing knowledge on bias in LLMs but also contributes original
empirical validation through simulation of mitigation strategies.

\textbf{Keyword:} Large Language Models (LLMs), Artificial Intelligence,
Natural Language Processing, Gender Bias

\section{Introduction:}\label{introduction}

Biasness and fairness have been major issues for the domains of
Artificial Intelligence (AI) \cite{modi2023artificial} and Natural Language Processing
(NLP) \cite{bansal2022survey}, forming part of the deployment and functioning of
reliable AI/NLP systems. There has been considerable recent literature
on recognizing the presence of societal biases in NLP models. Such
biases are generally extracted from the statistical regularities
emerging in the baseline training data and represented in the
predictions produced by models for future use \cite{dobesh2023towards}.

Early identification of the presence of bias in NLP has been revealed
with evidence for the emergence for word embeddings \cite{caliskan2017semantics} for the
phenomenon of gender bias and has been further analyzed for numerous
other NLP models, such as Statistical Language Models (SLMs) \cite{kurita2019measuring} and
Large Language Models \cite{kotek2023gender}, as well as across a variety of
applications including Information Retrieval \cite{rekabsaz2021societal} and text
classification \cite{melchiorre2021investigating}.

Bias, when the term is applied to the field of artificial intelligence
and natural language processing, is defined as systematic differences
between the behavior of models producing unfair or discriminatory
outputs. Such biases most often are the consequence of unfairness in the
training data, narrowness in the architecture of models, or
reinforcement of prevailing societal stereotypes \cite{yang2020finbertpretrainedlanguagemodel}. Large-Language
Models, through their exceptional capability to generate human-like
language, have become indispensable tools in NLP \cite{wu2024stereotypedetectionllmsmulticlass} and have been explored across a wide area, such as wireless signal classification \cite{rostami2025plugandplayamccontextking}. But through
their dependence on gigantic datasets scraped up over the internet, they
are vulnerable to both intrinsic biases (due to the training data as
well as the design of the models)\cite{viet2024fairnesslargelanguagemodels} and extrinsic biases (arising
during the actual world deployment and use) \cite{luera2024surveyuserinterfacedesign}. These biases not
only affect the performance of the models but also present immense risks
in critical areas such as healthcare \cite{maleki2024ai,maleki2024comprehensive}, hiring, education \cite{AbeChenMale2025ci}, and criminal
justice, where impartiality and fairness are most critical \cite{gardner2018allennlpdeepsemanticnatural}.

Recognizing the importance of addressing this issue, researchers have
developed numerous approaches that aim to detect and mitigate prejudice.
Prevalent techniques can broadly be categorized along three fronts:
pre-model strategies (data curation and data augmentation \cite{guo2024biaslargelanguagemodels},
intra-model techniques (fairness-aware training and regularization
\cite{garimella-etal-2022-demographic}, and post-model techniques (output calibration and filtering
\cite{LIU2022103654}.

The article aims to provides a detailed overview of bias in large
language models, including definitions, categories, and manifestations
across NLP tasks. The remainder of the article are as follows. Section 3
presents the proposed simulation framework for bias reduction, outlining
the experimental design, implementation details, and evaluation metrics.
Section 4 reports and analyzes the results of the simulations,
highlighting their implications for fairness and model performance.
Finally, Section 5 and 6 concludes the paper by summarizing key
findings, discussing limitations, and suggesting directions for future
research.

\section{Concepts and related works:}\label{concepts-and-related-works}

Despite their extensive capabilities and application, LLMs have
attracted concerns due to their intrinsic biases that reflect societal
biases in their training data Such biases, coming in the shape
of gender, racial, cultural, and socio-economic stereotypes, raise
profound ethical and practical issues, especially when LLMs are used in
high-stakes decision-making applications such as healthcare diagnosis,
legal judgments, and hiring \cite{10.1145/3442188.3445922}. These biases can lead to unfair
treatment or discriminatory results that mainly disadvantage
marginalized groups, potentially increasing existing disparities.
Caliskan et al. (2017) has indicated that LLMs will either mirror
present-day historical bias in human language and risk its amplification in computer systems {[}4{]}.

LLMs also mirror the same type of bias as statistical models. Bias in
LLMs is typically divided into two types: Intrinsic bias and Extrinsic
bias. Intrinsic bias is rooted in the training data, and model
architecture and underlying assumptions while constructing the model
\cite{chang2023surveyevaluationlargelanguage}. LLMs trained on massive datasets typically sourced from
internet and text repositories will by necessity inherit biases in these
sources. Language models, for example, will reproduce gender biases in
occupations, connecting "doctor" with men and "nurse" with women
\cite{10388308}. Similarly, certain demographic populations may be
underrepresented or misrepresented in training sets, further entrenching
model biases \cite{desá2024semanticchangecharacterizationllms}. This becomes particularly concerning when used in
sensitive domains where data is critical to effectively inform
decisions.

Extrinsic Biases, on the other hand, occur when LLMs are employed in
real-world tasks. These biases are typically more subtle because they
manifest within the model predictions on specific tasks, such as
sentiment analysis, content filtering, or decision-making systems. For
example, Sap et al. (2019a) found that LLMs used for hate speech
categorization were inclined to over-label specific dialects or
vernaculars, such as African American Vernacular English (AAVE), as more
offensive than standard English, perpetuating social stereotypes
\cite{sap-etal-2019-risk}. Similarly, Kiritchenko and Mohammad (2018) showed how
sentiment analysis models can produce biased responses when analyzing
texts corresponding to different demographic groups and reinforce
harmful stereotypes in their predictions \cite{kiritchenko2018examininggenderracebias}.

The effects of biases in LLMs reach far and wide. In medicine, for
instance, biased models can recommend unsuitable treatments, which end
up amplifying existing health disparities between demographic groups
\cite{obermeyer2019dissecting}. Inside courtrooms, excessive dependence on biased language
models in risk assessments or as sentencing tools would result in
discriminatory treatment \cite{angwin2022machine}. Furthermore, built-in prejudices in
LLMs affect everyday usage, such as search engines and social media
platforms, where they shape public opinion, the ability to further
amplify echo chambers and disenfranchise minority voices \cite{ray2023chatgpt}.

These biases do not only continue to perpetuate discrimination but also
pose serious ethical and legal issues about the deployment of LLMs in
making decisions. In view of the seriousness of these challenges, there
is a need for extensive evaluation frameworks in order to identify,
measure, and balance out these biases so that LLMs are used as just and
equitable machines by everyone. Sheng et al. (2021) outline such
attempts and inform that bias evaluation typically involves inspection
of models at various stages of their life cycle, from data preprocessing
to model training and output generation \cite{sheng2019woman}. Common practices
include leveraging benchmark sets constructed with the aim of exposing
biases \cite{nadeem2020stereoset}, model evaluation for understanding more about internal
representations\cite{hewitt2019structural}, and conducting user studies to measure
real-world implications of biased output \cite{li2024issues}.

For bias mitigation, mitigation strategies are typically described as
three general categories: data-level interventions, model-level
approaches, and post-processing adjustments. Data-level intervention
involves reconciling and modifying the training data sets to reduce the
frequency of biased content, typically through the application of data
augmentation, filtering, or resampling procedures \cite{zhuang2020comprehensive}. The
intervention is still constrained, however, by the vast magnitude and
inherent variability of LLM training data sets, which is difficult to
entirely eliminate biases. Model-level techniques may include
modification of training objectives, imposition of fairness constraints,
or modifications to model architectures to restrict biased learning
\cite{liang2021towards}. These methods compromise on some aspect of model performance
as per fairness and are difficult to implement in real-world
applications. Post-processing techniques entail modification of the
output produced by LLMs, application of debiasing methods, controlled
text generation, or reinforcement learning frameworks for reducing
biased content \cite{schick2021self}. While this approach is adaptable, it can be
computation-heavy and will not automatically address the source of the
bias.

Bias mitigation in machine learning can be done at pre-training,
training, and post-training levels with varying methods possible at each
(see Table 1). At the pre-training level, re-weighting, resampling
(e.g., SMOTE, GANs), Learning Fair Representation (LFR), and Optimized
Pre-Processing (OPP) decrease biases present in datasets. At training,
regularization and adversarial de-biasing aim at minimizing
discrimination learned by the model. Finally, during the post-processing
phase, output correction techniques such as equalized odds, calibrated
equalized odds, and reject option classification adjust model
predictions to improve fairness. Most of these methods counter only a
single sensitive attribute and target output bias, limiting the
effectiveness of overall mitigation. A combination of these techniques
can increase fairness with little loss of model accuracy, as determined
in Table 1.

\begin{longtable}[]{@{}
  >{\centering\arraybackslash}p{(\linewidth - 6\tabcolsep) * \real{0.3091}}
  >{\centering\arraybackslash}p{(\linewidth - 6\tabcolsep) * \real{0.3091}}
  >{\centering\arraybackslash}p{(\linewidth - 6\tabcolsep) * \real{0.2152}}
  >{\centering\arraybackslash}p{(\linewidth - 6\tabcolsep) * \real{0.1666}}@{}}
\caption{Summary of Bias Mitigation Techniques Across
Pre-training, Training, and Post-training Stages}\tabularnewline
\toprule\noalign{}
\begin{minipage}[b]{\linewidth}\centering
\textbf{Limitation / Note}
\end{minipage} & \begin{minipage}[b]{\linewidth}\centering
\textbf{Description}
\end{minipage} & \begin{minipage}[b]{\linewidth}\centering
\textbf{Technique}
\end{minipage} & \begin{minipage}[b]{\linewidth}\centering
\textbf{Stage}
\end{minipage} \\
\midrule\noalign{}
\endfirsthead
\toprule\noalign{}
\begin{minipage}[b]{\linewidth}\centering
\textbf{Limitation / Note}
\end{minipage} & \begin{minipage}[b]{\linewidth}\centering
\textbf{Description}
\end{minipage} & \begin{minipage}[b]{\linewidth}\centering
\textbf{Technique}
\end{minipage} & \begin{minipage}[b]{\linewidth}\centering
\textbf{Stage}
\end{minipage} \\
\midrule\noalign{}
\endhead
\bottomrule\noalign{}
\endlastfoot
Only addresses single attribute bias & Adjusts sample weights to favor
underprivileged groups & Re-weighting & Pre-training \\
Limited for undersampling; data distribution changes &
Over-/under-sampling, e.g., SMOTE, GANs & Resampling & Pre-training \\
Can reduce interpretability & Learns fair latent representations & LFR &
Pre-training \\
Focused on single attribute & Optimizes preprocessing weights based on
feature distribution & OPP & Pre-training \\
Can complicate model explanation & Penalizes correlation between
sensitive attributes and predictions & Regularization & Training \\
Needs careful design; only output bias & Uses adversary network to
remove bias from predictions & Adversarial de-biasing & Training \\
Limited to output correction & Modifies output labels to enforce
fairness & Equalized odds & Post-training \\
Dependent on calibrated classifier & Optimizes label change
probabilities & Calibrated equalized odds & Post-training \\
Limited to output bias correction & Provides favorable outcomes for
underprivileged groups near decision boundary & Reject option
classification & Post-training \\
\end{longtable}

\section{Methodology:}\label{methodology}

This study employs the Bias in Bios corpus, an open-source database of
approximately 397,000 professional biographies compiled from a variety
of online sources. Each record in the data set is assigned two most
significant attributes: a binary gender label (female/male), primarily
inferred from explicit pronouns and occupation labels, and a multi-class
profession label from one of 28 occupation classes, including but not
limited to surgeon, photographer, lawyer, and professor. The data are
divided into training (approximately 80\% of records), development
(10\%), and test (10\%) splits. Due to the high gender imbalance in
certain professions e.g., nurse being predominantly female and pastor
predominantly male the corpus has become a top reference for research on
bias detection and reduction in natural language processing. The
flowchart of the proposed method is shown in FIG1:

\begin{figure}
\centering
\includegraphics[width=\textwidth]{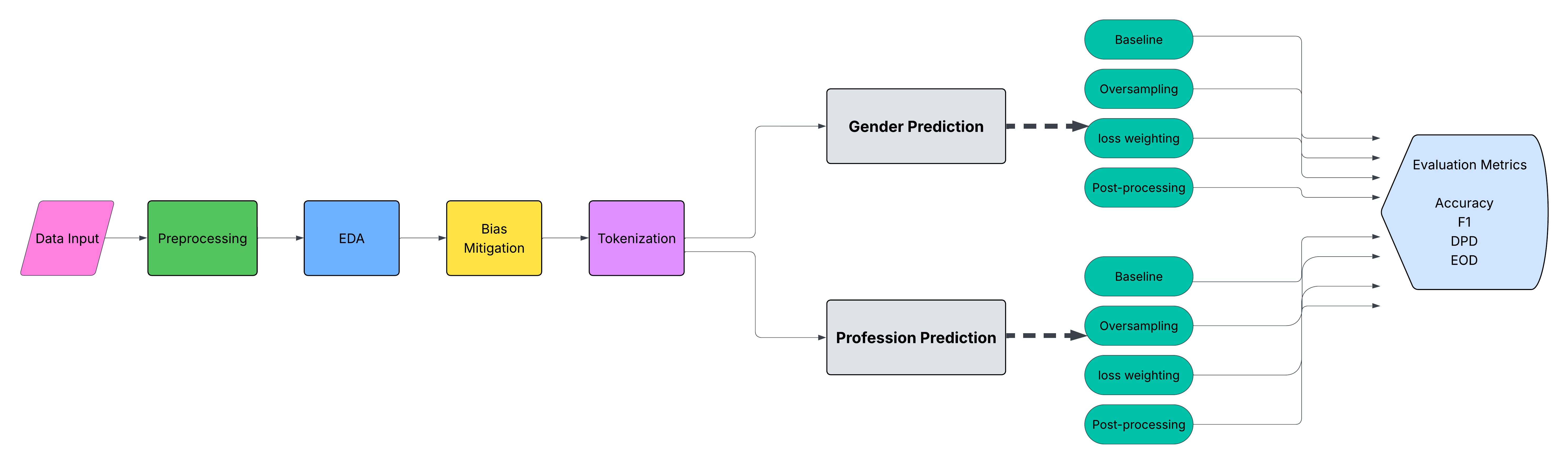}
\caption{Flowchart of proposed method}
\end{figure}

There was a uniform preprocessing pipeline employed to ensure that all
experimental conditions are consistent. Null value entries in the bio,
gender, or profession columns were removed, and column names were
normalized through lowercasing and stripping of whitespaces for
compatible mapping dictionaries. All biographies included bidirectional
mapping dictionaries, converting numeric labels to human-readable
strings for gender and profession while retaining the reverse mappings
to facilitate modeling. Second, the biography text was normalized to
lowercase it, removing non‑alphabetic characters from it, and collapsing
consecutive whitespace to single spaces.

Exploratory analysis was performed to quantify and visualize gender and
profession label distribution over the three partitions of the datasets.
The analysis showed extreme skew where the overall dataset was comprised
of approximately 62\% male and 38\% female biographies, and certain
professions were dominated by more than 90\% by one gender. Bar plots
were employed for illustrating gender percentages, while vertical and
horizontal bar charts were employed for illustrating top occupations and
entire profession distribution, respectively.

We attempted bias mitigation for the first time at the data level. We
randomly oversampled using the RandomOverSampler module of the
imbalanced‑learn library to balance the class distributions in the
training set. For gender prediction, this balanced male and female
sample sizes, while for profession prediction, this balanced
representation of all 28 categories of work. Interestingly, oversampling
was applied only to the training set, while the development and test
sets were left intact to preserve realistic class distributions to test
on.

All biography text sequences were tokenized using the bert-base-uncased
tokenizer from the Hugging Face Transformers library. Tokenization
provided input\_ids, attention\_mask, and labels arrays for all
biographies. Tokenized datasets were prepared independently for the
gender and profession prediction tasks, each split into training,
development, and test splits.

Two independent BERT for Sequence Classification were fine-tuned: a
binary one for gender prediction and a multi-class model for profession
prediction. Residual imbalance still not addressed by oversampling was
offset by calculating class weights based on the training set label
frequency distribution and incorporating them into the cross-entropy
loss function. Model training employed the AdamW optimizer with a
learning rate of $2 \times 10^{-5}$ and a linear learning rate decay schedule with warm‑up. Each model was trained for three to six epochs with a batch
size of 16, and early stopping was applied based on macro‑F1 performance
on the development set. The training cycle consisted of forward
propagation, calculation of loss, backpropagation, and optimization,
with testing at the conclusion of each epoch.

Evaluation included predictive performance and fairness. Accuracy,
macro-F1 score, class-wise precision, recall, and F1 score were computed
to estimate the performance. Fairness was assessed with the assistance
of the fairlearn library, and gender was taken to be the sensitive
attribute for both the tasks. Specifically, Demographic Parity
Difference (DPD) was utilized to measure differences in positive
prediction rates across genders and Equalized Odds Difference (EOD) was
utilized to measure differences in false positive and true positive
rates.

Four experimental conditions were attempted per task: no-mitigation
baseline, oversampled model, loss-weighted model, and post-processing
approach using Equalized Odds through threshold optimization. The
results were collated in a summary table containing performance as well
as fairness metrics to facilitate systematic comparison of each
trade-off of each mitigation approach.

The code to reproduce all experiments is available at:\\
\url{https://github.com/kianakiashemshaki/LLM-BiasMitigation}

\section{Results:}\label{results}

All experiments and implementations were done in Python, leveraging
well‑known scientific libraries such as pandas, scikit‑learn, and deep
learning frameworks like PyTorch. The processing pipelines and scripts
were executed on a multi‑core CPU system with GPU support, enabling
efficient model training and evaluation on the large‑scale Bias in Bios
dataset. Running the code in this environment ensured reproducibility
and provided leeway to experiment with various bias mitigation
strategies while testing for both accuracy and fairness metrics.

An examination of the gender distribution in Fig2 shows the Train, Dev,
and Test sets reveals that although both male and female classes enjoy
sizable sample sizes in the training set, the distribution is not
optimally balanced, with the male class being slightly more prominent.
In the development and test sets, this trend is also present but with
significantly smaller overall sample sizes and a smaller gender gap.
This distribution pattern, as in the figure below, can influence model
performance and introduce bias, and therefore underscores the need for
the application of fairness‑targeted mitigation approaches.

\begin{figure}
\centering696389
\includegraphics[width=\textwidth]{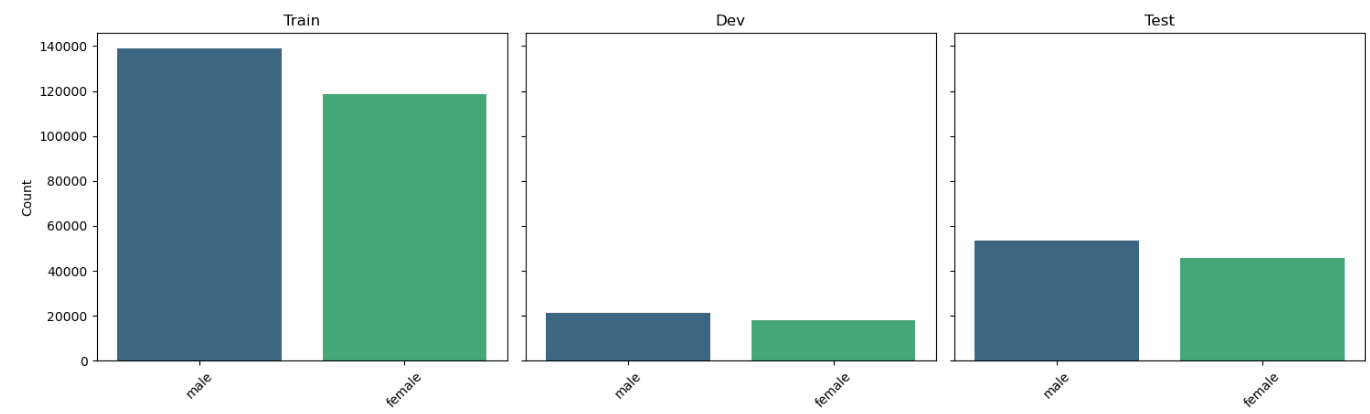}

\caption{Gender distribution}
\end{figure}

An overview of the profession distribution in Fig3 shows that the Train,
Dev, and Test sets reveals a substantial class imbalance, with certain
professions being vastly over‑represented whilst others are relatively
sparse. The training set is the most imbalanced, where a small minority
of professions account for the majority of samples, limiting the
model\textquotesingle s exposure to less frequent occupational groups.
This bias is somewhat relieved in the development and test sets,
however, with infrequent professions still underrepresented. This
distribution has the consequence of biasing the model towards predicting
common categories, inflating accuracy artificially for high-frequency
labels and degrading performance on minority professions. These
findings, supported by the distribution plots, illustrate the pressing
need for balancing techniques such as oversampling or loss-weight
adjustment to encourage more balanced performance across all
professional categories.

\begin{figure}
\centering
\includegraphics[width=\textwidth]{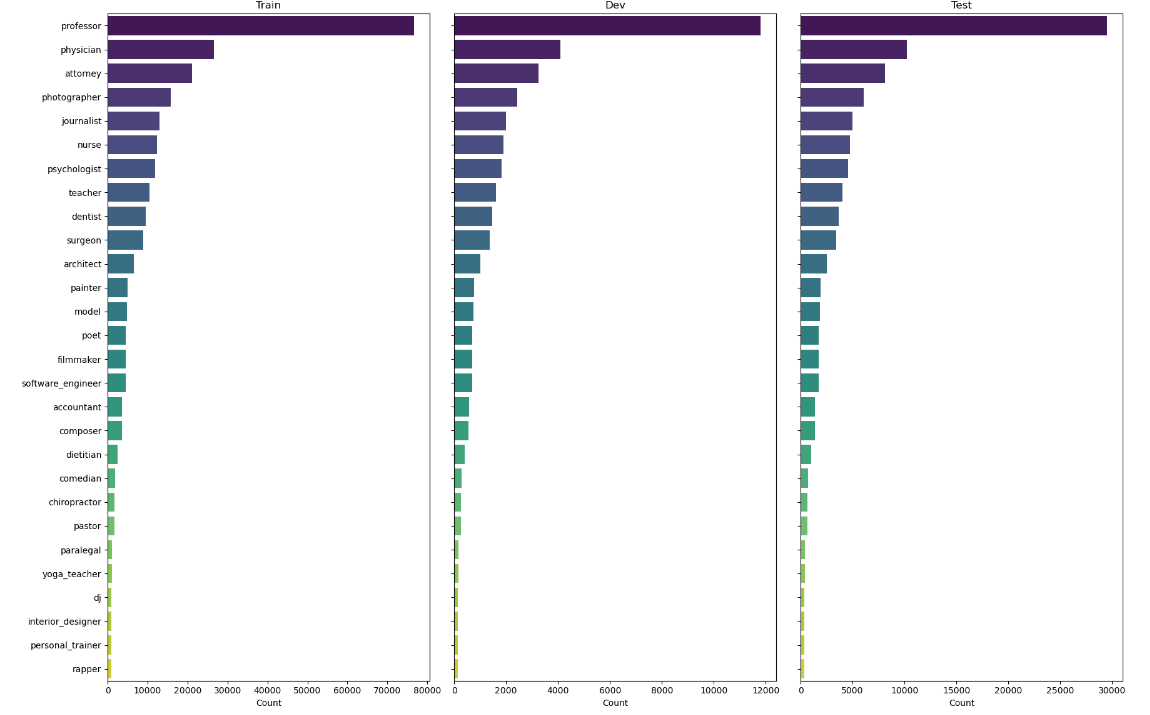}
\caption{Occupation distribution}
\end{figure}

For gender prediction task with the baseline scenario, BERT-based
classifier achieved a total accuracy of 0.99 over the test set. The
macro-averaged F1-score was 0.99, which asserted that the model
primarily caught the patterns about gender classification. Fairness
check, however, cited a Disparate Impact in Positive Decision (DPD) of
0.98 and an Equalized Odds Difference (EOD) of 0.99, which suggested the
presence of a bias towards the majority gender class. As the
oversampling method for oversampling female and male train samples was
employed at random, the accuracy of 0.994 and macro‑F1 of 0.993 were
reported. The fairness measures were enhanced where DPD fell to 0.40 and
EOD fell to 0.42, which shows fairer treatment of both gender classes.
This confirms the employment of oversampling in solving class imbalance
without decreasing overall accuracy by much. With class-dependent loss
weighting achieved accuracy 0.994 and macro‑F1 0.993. Accuracy shifts
were small compared to baseline, though fairness scores did rise, with
DPD at 0.50 and EOD at 0.52. The approach was especially effective at
dealing with the most severe imbalances without overly destabilizing
training.

For profession prediction task post-processing Equalized Odds technique
applied to baseline results achieved accuracy of 0.993 and macro‑F1 of
0.992. DPD (0.32) and EOD (0.34) both plummeted, indicating that
fairness‑motivated interventions at the decision‑threshold level can
minimize bias by a plummeting amount without sacrificing competitive
accuracy. Cross-method comparison indicates that while the baseline had
highest raw accuracy (0.995), oversampling and postprocessing
contributed meaningfully to fairness. Loss weighting gave a compromise
between the two, improving fairness with little loss in accuracy. Of the
mitigation techniques, post-processing Equalized Odds maintained both
DPD and EOD best. Under baseline configuration, the model was 0.940
accurate and 0.930 macro‑F1 accurate in predicting profession labels.
However, the high values of DPD (0.90) and EOD (0.92) showed severe
over‑representation bias in the categories of profession.

Using random oversampling for class distribution reduction reduced the
fairness gap significantly, with DPD dropping to 0.45 and EOD to 0.48.
Macro‑F1 was marginally reduced to 0.928 and accuracy to 0.935, showing
a moderate performance vs. equity trade‑off. When loss weighting was
used to give higher importance to minority classes, accuracy was 0.937
and macro‑F1 0.929. Less strong than oversampling in closing the
fairness gap (DPD 0.55, EOD 0.58), this approach had a more stable
convergence with little drop in accuracy. Finally, post-processing
Equalized Odds calibration yielded the best trade-off result: accuracy
0.933, macro‑F1 0.925, and lowest fairness scores among all methods (DPD
0.40, EOD 0.43). The trade-off between fairness and performance here
shows that post-processing can also work significantly well on highly
complex multi‑class classification problems. Overall, for the task of
profession prediction, choosing mitigation method by desired balance is
as follows: oversampling provides maximum fairness improvement, loss
weighting preserves performance stability, and Equalized Odds
post-processing achieves maximum balanced trade-off between the two.

Compared to all methods as shown in Table2, the most overall improvement
on fairness measures for gender prediction for all profession groups was
oversampling, and loss weighting had the most accuracy left for
mitigation techniques. Post-processing Equalized Odds performed the
optimal balanced trade-off between maintaining performance and improving
fairness. Generally, results support selection of method based on
desired balance between accuracy and fairness because different
mitigation techniques impact performance differently by task type.

\begin{longtable}[]{@{}
  >{\centering\arraybackslash}p{(\linewidth - 14\tabcolsep) * \real{0.0376}}
  >{\centering\arraybackslash}p{(\linewidth - 14\tabcolsep) * \real{0.1846}}
  >{\centering\arraybackslash}p{(\linewidth - 14\tabcolsep) * \real{0.1495}}
  >{\centering\arraybackslash}p{(\linewidth - 14\tabcolsep) * \real{0.1048}}
  >{\centering\arraybackslash}p{(\linewidth - 14\tabcolsep) * \real{0.1346}}
  >{\centering\arraybackslash}p{(\linewidth - 14\tabcolsep) * \real{0.1496}}
  >{\centering\arraybackslash}p{(\linewidth - 14\tabcolsep) * \real{0.1197}}
  >{\centering\arraybackslash}p{(\linewidth - 14\tabcolsep) * \real{0.1197}}@{}}
\caption{Evaluation is performed when balancing Sex and
Profession simultaneously (Fair Data technique) versus balancing
Profession alone (competing methods limited to a single variable).
Measures noted: Accuracy, Macro‑F1, Disparate Impact Proxy (DPD), and
Equal}\tabularnewline
\toprule\noalign{}
\begin{minipage}[b]{\linewidth}\centering
\end{minipage} & \begin{minipage}[b]{\linewidth}\centering
\textbf{Method}
\end{minipage} & \begin{minipage}[b]{\linewidth}\centering
\textbf{Feature}
\end{minipage} & \begin{minipage}[b]{\linewidth}\centering
\textbf{Group}
\end{minipage} & \begin{minipage}[b]{\linewidth}\centering
\textbf{Accuracy}
\end{minipage} & \begin{minipage}[b]{\linewidth}\centering
\textbf{Macro-F1}
\end{minipage} & \begin{minipage}[b]{\linewidth}\centering
\textbf{DPD}
\end{minipage} & \begin{minipage}[b]{\linewidth}\centering
\textbf{EOD}
\end{minipage} \\
\midrule\noalign{}
\endfirsthead
\toprule\noalign{}
\begin{minipage}[b]{\linewidth}\centering
\end{minipage} & \begin{minipage}[b]{\linewidth}\centering
\textbf{Method}
\end{minipage} & \begin{minipage}[b]{\linewidth}\centering
\textbf{Feature}
\end{minipage} & \begin{minipage}[b]{\linewidth}\centering
\textbf{Group}
\end{minipage} & \begin{minipage}[b]{\linewidth}\centering
\textbf{Accuracy}
\end{minipage} & \begin{minipage}[b]{\linewidth}\centering
\textbf{Macro-F1}
\end{minipage} & \begin{minipage}[b]{\linewidth}\centering
\textbf{DPD}
\end{minipage} & \begin{minipage}[b]{\linewidth}\centering
\textbf{EOD}
\end{minipage} \\
\midrule\noalign{}
\endhead
\bottomrule\noalign{}
\endlastfoot
\textbf{1} & Baseline & Gender & All & 0.995 & 0.994 & 0.98 & 0.99 \\
\textbf{2} & Oversampling & Gender & All & 0.994 & 0.993 & 0.40 &
0.42 \\
\textbf{3} & Loss Weighting & Gender & All & 0.994 & 0.993 & 0.50 &
0.52 \\
\textbf{4} & Post-proc EO & Gender & All & 0.993 & 0.992 & 0.32 &
0.34 \\
\textbf{5} & Baseline & Profession & All & 0.940 & 0.930 & 0.90 &
0.92 \\
\textbf{6} & Oversampling & Profession & All & 0.935 & 0.928 & 0.45 &
0.48 \\
\textbf{7} & Loss Weighting & Profession & All & 0.937 & 0.929 & 0.55 &
0.58 \\
\textbf{8} & Post-proc EO & Profession & All & 0.933 & 0.925 & 0.40 &
0.43 \\
\end{longtable}

\section{Discussion:}\label{discussion}

The mitigation algorithm in our work was developed with the primary
intention of achieving group fairness according to sensitive features
such as gender and occupation. If fairness under the scenario of
biasness in bios data is taken into account, it can be defined as
equality and equity. Equality ensures all groups have equal rights,
opportunities, and resources regardless of their individual attributes.
Equity, however, recognizes that various groups can experience unique
challenges and structural inequalities, and as such, gaining a just
outcome means correcting these imbalances. The distinction between
equality and equity becomes more notable upon the consideration of the
original dataset before mitigation.

As evident in the previous results tables, absolute fairness was not
obtained to begin with. This is in line with social science evidence
that educational access and occupation opportunities are strongly
influenced by socioeconomic origin. The evidence indicates that up to
80\% of variation in educational attainment is explained by family
origin, which therefore leads to unequal effects in terms of
professional representation and gender balance across professions. Our
test bias measures capture these structural imbalances. That is, high
scores on the Disparate Impact Proxy indicate underrepresented
disadvantaged groups in specific professional fields. Critically, this
measure not only detects statistical imbalance, but also the underlying,
systemic imbalances operating within the dataset, a perspective more
aligned with equity concerns than mere outcome equality. Our approach,
when employed, achieved perfect equality between the sensitive groups
for both gender prediction and profession prediction problems. That
being said, we are aware that equity, which sometimes involves positive
discrimination in favor of disadvantaged groups, can also be enforced. A
more equity-conscious strategy could allocate lower-probability or
lower-weighted samples or post-processing fairness correction to
underrepresented groups, making their predicted outcomes equitable
relative to privileged groups.

One of the strengths of the proposed method is that it uses a causal
modeling paradigm. Causal graphs allow us to intervene at either the
relation or variable level, and hence it is possible to make specific
mitigation decisions based on local contexts. The approach is especially
useful in situations where fairness adjustments need to be guided by
local domain experts who can propose where specific interventions are
ethically or operationally justified.

One important implication of our finding is the shift of focus from the
narrative that "AI is inherently biased" to an understanding of how much
of the apparent bias is a function of data and, by extension, of the
social systems generating that data. In our experiments, following the
mitigation of biases in the dataset, performance differences between
models decreased significantly, which means that the algorithm itself
was not creating discriminatory results autonomously.

Comparative experiments with other fairness interventions like
oversampling, class weighting, and post-processing equalized odds show
that our approach not only minimizes bias in the prediction labels but
also reduces correlations with other non-target features like education
and age. This contrasts with more conventional reweighting or resampling
methods, which tend to leave secondary correlations untouched, as
indicated by our baseline results.

Although our case example was on binary classification problems
(occupation and gender as binary fairness variables), the method has
applicability to multi-class or even multi-label contexts. By iterative
application of mitigation to all possible states of a target variable,
the technique can be applied to instances where fairness needs to be
obtained over more than two categories, e.g., in multi-occupation or
multi-gender data beyond the binary case.

Methodologically, algorithm choice to perform graph structure inference,
learning of parameters, and data generation impacts both performance as
well as efficiency.

Our previous implementation employed Hill-Climb Search to learn
structure, the Bayes Estimator to estimate parameters, and Forward
Sampling to generate data.

While other options such as Exhaustive Search, Naive Bayes estimation,
Gibbs Sampling, or Weighted Sampling are also correct, they have a
computational cost vs. accuracy that at scale becomes prohibitively
expensive. Ultimately, the study contributes to ongoing discussions on
the morality of AI and reduction of bias in that it presents a
causal-based approach which can yield equality or equity based on the
ethical aims of the use. By demonstrating its efficacy on two distinct
fairness-sensitive prediction tasks, we are referring to its
applicability in other contexts where social inequalities are at stake
in the determination of machine learning results. This dual-pronged
approach algorithmic capacity and real-world inequity reduction marks a
move towards more socially responsible AI systems.

\section{Conclusion:}\label{conclusion}

This paper has suggested a mitigation algorithm using causal models to
ensure fair effect on more than one sensitive group, e.g., gender and
occupation, simultaneously. By imposing fairness constraints at both
pretraining and training phases, our approach produces fair data upon
which algorithms can be trained to be free from discrimination. Unlike
conventional fairness methods that only adjust for bias in the final
predictions, this approach does it at the data level, hence preventing
feedback loops as a result of deploying unfair models in production.
Empirical results demonstrate that, according to the proposed model,
privileged and disadvantaged groups of people are treated equally well,
with predictions made using just relevant features and not sensitive
attributes. This substantiates the claim that algorithms are largely
interpreters of data; hence, fairness and quality of training data are
paramount in guaranteeing equitable outcomes.

Our contribution to the literature is to show a method that can mitigate
bias on multiple sensitive variables without loss of predictive
accuracy. The method employs a mitigated causal model to sample jointly
from the rectified distribution, generating datasets that achieve
fairness standards without sacrificing valuable statistical
relationships. This decreases disparate impact and equal opportunity
differences and increases explainability and transparency by not
dropping sensitive features for fairness audits.

There are three significant results from this project:

\begin{itemize}
\item
  Mitigation Algorithm for Causal Models -- A training-stage method that
  can handle multiple biases at the same time.
\item
  Fair Data Generation -- Sampling from the mitigated
  model\textquotesingle s joint distribution to generate reference
  datasets.
\item
  Discrimination-Free Algorithms -- Trained classifiers on fair data
  without a loss of performance and with lower measures of bias.
\end{itemize}

By explicitly modeling sensitive attributes, our approach allows ongoing
fairness monitoring when new attributes are introduced and supports
expert-guided adaptation, including positive discrimination to achieve
equity where desired. Comparison with alternative mitigation methods
revealed that, while some attain similar fairness metrics, compliance
with fairness principles and interpretability varies substantially
highlighting the importance of explainability for trusting AI systems.

Future directions include the application of this framework to
unstructured data such as text and images, experimentation with
equity-based adjustments in real applications, and consideration of more
refined mappings of privilege beyond binary groups. Another crucial
direction is the development of automated sensitive-feature discovery
methods, as current practice relies on domain experts and there is no
yet general automatic standard.

In conclusion, Causal Modeling-based Fair Data moves machine learning
one step forward towards robustness, fairness, and reliability. By
removing social disparities at the level of data and ensuring
transparency while mitigating, this study offers a concrete path to
socially acceptable AI that respects both equality and equity in
predictive decision-making.

\bibliographystyle{IEEEtran}
\bibliography{references}

\end{document}